\documentclass[conference]{IEEEtran}
\IEEEoverridecommandlockouts
\usepackage{cite}
\usepackage{amsmath,amssymb,amsfonts}
\usepackage{algorithmic}
\usepackage{graphicx}
\usepackage{textcomp}
\usepackage{xcolor}
\usepackage{booktabs}
\usepackage{hyperref}
\def\BibTeX{{\rm B\kern-.05em{\sc i\kern-.025em b}\kern-.08em
    T\kern-.1667em\lower.7ex\hbox{E}\kern-.125emX}}
\begin{document}
\title{SelaFD:Seamless Adaptation of Vision Transformer Fine-tuning for Radar-based Human Activity Recognition
\\
\thanks{The work was supported by the National Natural Science Foundation of China under Grants No. 12374421, and in part by the Key Laboratory of Underwater Acoustic Countermeasure Technology under Grant No. 2022JCJQLB03305. Corresponding Email:qisong.wu@seu.edu.cn.

This work has been submitted to the IEEE for possible publication. Copyright may be transferred without notice, after which this version may no longer be accessible.}
}

\author{
\IEEEauthorblockN{Yijun Wang\IEEEauthorrefmark{2}$^{1}$, Yong Wang\IEEEauthorrefmark{2}$^{2}$, Chendong Xu\IEEEauthorrefmark{2}$^{3}$, Shuai Yao\IEEEauthorrefmark{2}$^{4}$, Qisong Wu\IEEEauthorrefmark{2}\IEEEauthorrefmark{3}\IEEEauthorrefmark{1}}
\IEEEauthorblockA{\IEEEauthorrefmark{2}
    Key Laboratory of Underwater Acoustic Signal Processing of Ministry of Education, \\ Southeast University, Nanjing, 210096, China \\
\IEEEauthorblockA{\IEEEauthorrefmark{3}
    Purple Mountain Laboratories, Nanjing, 211111, China}
}
\IEEEauthorblockA{
$^{1}$wang-yj@seu.edu.cn
$^{2}$yong.wang@seu.edu.cn
$^{3}$230228233@seu.edu.cn
$^{4}$shuaiyaoseu@seu.edu.cn
\IEEEauthorrefmark{1}qisong.wu@seu.edu.cn
}
}

\maketitle
\begin{abstract}
Human Activity Recognition (HAR) such as fall detection has become increasingly critical due to the aging population, necessitating effective monitoring systems to prevent serious injuries and fatalities associated with falls. This study focuses on fine-tuning the Vision Transformer (ViT) model specifically for HAR using radar-based Time-Doppler signatures. Unlike traditional image datasets, these signals present unique challenges due to their non-visual nature and the high degree of similarity among various activities. Directly fine-tuning the ViT with all parameters proves suboptimal for this application. To address this challenge, we propose a novel approach that employs Low-Rank Adaptation (LoRA) fine-tuning in the weight space to facilitate knowledge transfer from pre-trained ViT models. Additionally, to extract fine-grained features, we enhance feature representation through the integration of a serial-parallel adapter in the feature space. Our innovative joint fine-tuning method, tailored for radar-based Time-Doppler signatures, significantly improves HAR accuracy, surpassing existing state-of-the-art methodologies in this domain. Our code is released at \href{https://github.com/wangyijunlyy/SelaFD}{https://github.com/wangyijunlyy/SelaFD}.
\end{abstract}
\section{Introduction}
\label{sec:intro}

As the global population ages, the field of Human Activity Recognition (HAR) such as fall detection task has emerged as a critical area of research and application, particularly for the elderly. The World Health Organization reports that falls are the second leading cause of unintentional injury deaths, with the highest mortality rates occurring among individuals over 60 years old \cite{WHO2021}. This alarming statistic underscores the urgent need for effective monitoring systems that can mitigate the risks associated with falls, thereby improving the quality of life for older adults and reducing healthcare costs and societal burdens.

Traditional methods for e.g. fall detection tasks in the HAR field, such as wearable devices \cite{Lee2015,wang2017low,lara2012survey,galvao2021multimodal} and vision-based systems \cite{Yu2012,shu2021eight,ray2023transfer}, have demonstrated effectiveness in certain scenarios; however, they come with significant limitations. Wearable devices may fail due to improper usage or discomfort, leading to non-compliance among users. Vision-based systems, on the other hand, often struggle in low-light conditions or complex environments where visibility is compromised. Furthermore, these methods frequently raise privacy concerns, as they rely on continuous monitoring of individuals in their living spaces, which may deter users from adopting such technologies.

Millimeter-wave radar technology presents a promising alternative for HAR, overcoming many of the shortcomings associated with traditional technologies. This technology operates effectively under various lighting conditions and can penetrate clothing to capture subtle motion changes. Additionally, the high resolution and directionality of millimeter-wave radar enable precise recognition of different activity patterns, significantly enhancing the accuracy of HAR systems \cite{Huang2020}.

With the rapid advancements in deep learning, particularly in the field of computer vision, researchers have begun to explore its applications in radar-based HAR.  Several efforts have been afforded to the HAR application, and many progresses have been acquired. 
A combination of One-Dimensional Convolutional Neural Networks (1DCNN) and Long Short-Term Memory (LSTM) to process radar spectrograms, capturing spatio-temporal properties effectively\cite{Zhu2020}.  
Multi-domain information fusion strategies for precise recognition of spatial micromotion targets\cite{Tian2022}. 
Lightweight deep convolutional networks like DIAT-RadHARNet, maintaining performance under adverse conditions\cite{Chakraborty2022}. 

All these studies above have been developed based on CNN and LSTM networks. Currently, the Vision Transformer(ViT) model has demonstrated remarkable capabilities in processing image data in various computer vision tasks\cite{Radford2021}, making it a compelling candidate for radar image analysis. Despite the success of pre-trained visual models, their effectiveness on radar-based images remains underexplored. The challenge lies in ensuring that the representations learned from natural images can be effectively transferred to radar images without significant loss of previously acquired knowledge, a problem known as catastrophic forgetting\cite{french1999catastrophic,kirkpatrick2017overcoming,luo2023empirical}. 

To tackle this challenge, we propose a novel approach called SelaFD (\textbf{Se}amless \textbf{L}earning for \textbf{A}daptive \textbf{F}all \textbf{D}etection), inspired by the concept of Parameter-Efficient Fine-Tuning (PEFT). While PEFT presents a promising solution, it has not yet been thoroughly explored in the context of radar-based HAR.
By using Low-Rank Adaptation (LoRA) fine-tuning\cite{hu2021lora} in the weight space and 
a serial-parallel adapter fine-tuning\cite{Houlsby2019} in the feature space into a pre-trained frozen model, we achieve 96.61\% classification accuracy on the University of Glasgow's publicly available dataset. Our contributions can be concluded as follows:

1. We first propose fine-tuned transfer learning using a pre-trained ViT to successfully apply the pre-trained knowledge to radar-based time-frequency image classification in the field of HAR.

2. We propose a novel joint fine-tuning method called SelaFD for the high similarity of Time-Doppler maps based on features of different granularity, which operates in both weight space and feature space for extracting the underlying coarse-grained features and the fine-grained features for improving the recognition of highly confusable categories.

3. Our proposed algorithms can be seamlessly connected to any ViT-based visual large-scale model or multimodal large-scale model, e.g., DINO\cite{caron2021emerging,oquab2023dinov2}, LLaVA\cite{liu2023llava,liu2023improvedllava,liu2024llavanext}. With the feature characterization capability of the larger-scale model, we can achieve higher accuracy in HAR.

\section{METHODS}
\label{sec:methods}
\begin{figure}[t]
    \centering
    \includegraphics[width=1\linewidth]{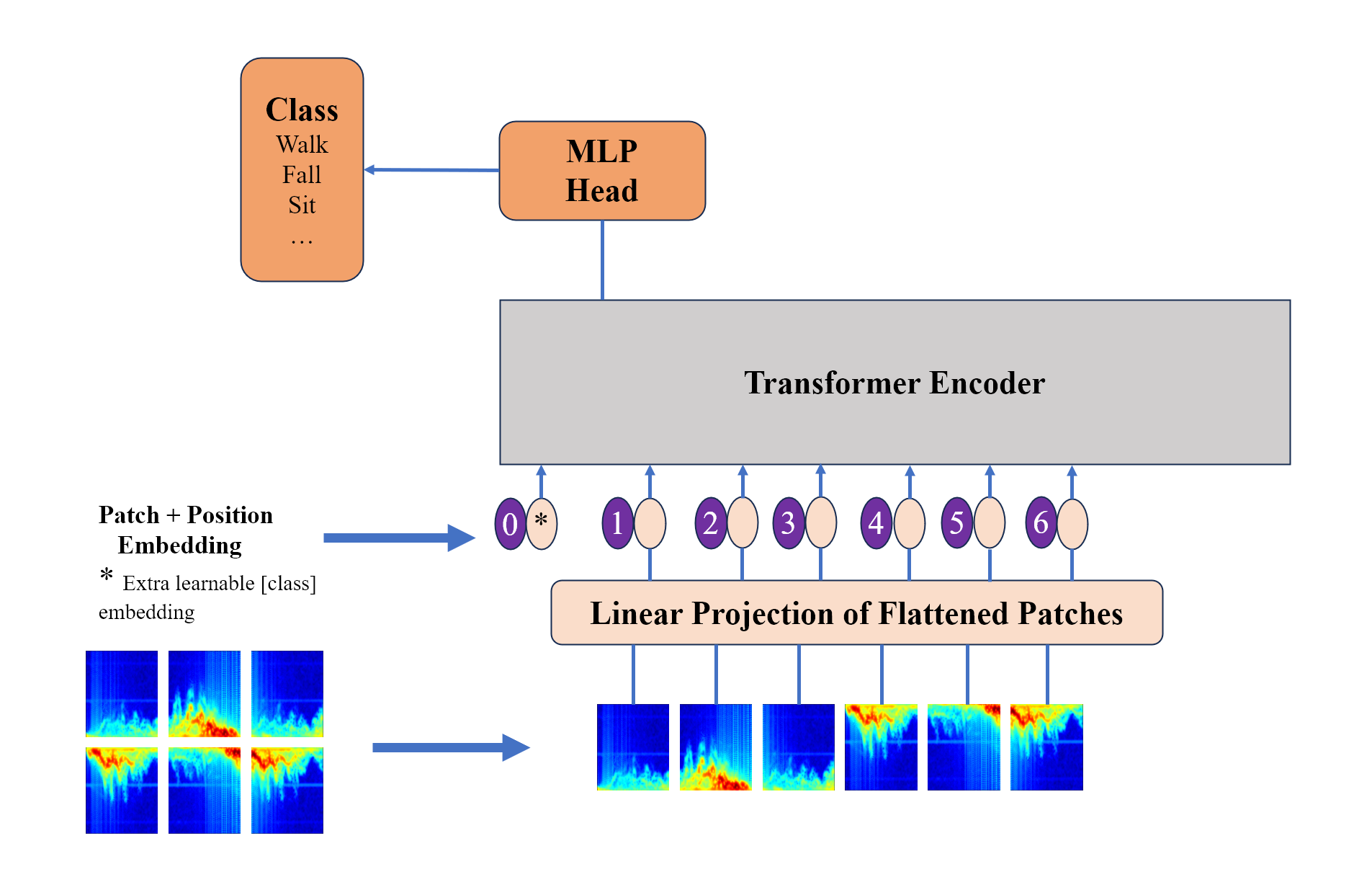}
    \caption{Architecture diagram of ViT.}
    \label{fig:vit}
\end{figure}

The architecture diagram of ViT in the radar-based HAR is shown in Fig. 1: the radar Time-Doppler (TD) spectrograms as inputs are added to the classical transformer encoder by patch embedding and position embedding, and finally an Multilayer Perceptron (MLP) mapping is added to complete our HAR task also known as human behavior classification. Due to the ViT's poor performance on radar images when used directly, we have fine-tuned the module mainly for the base transformer block in the transformer encoder. We first introduce LoRA, a widely-used weight-space fine-tuning method, to effectively transfer knowledge and extract basic coarse-grained features like shape, color and so on. 

Considering the differences between radar spectrum images and natural images i.e., highly confusable and strong similarity, such as the high similarity between two human activities, namely, drinking and picking up, in Fig. 3, we have demonstrated in the ablation experiment section that the only use of LoRA fine-tuning alone is still challenging in recognizing such highly similar human activities, and thus we need to further extract the fine-grained features on the basis of LoRA fine-tuning of the pre-trained ViT model. Therefore, to enhance the fine-grained feature representation, we introduce an additional serial parallel adapter module\cite{chen2022adaptformer,yang2023aim} in the feature space. As shown in Fig. 2, this section describes the SelaFD method for HAR, which combines the fine-tuning of the weight space and the feature space corresponding to the coarse- and fine-grained features of the spectrogram, respectively.
\begin{figure}[t]
    \centering
    \includegraphics[width=1\linewidth]{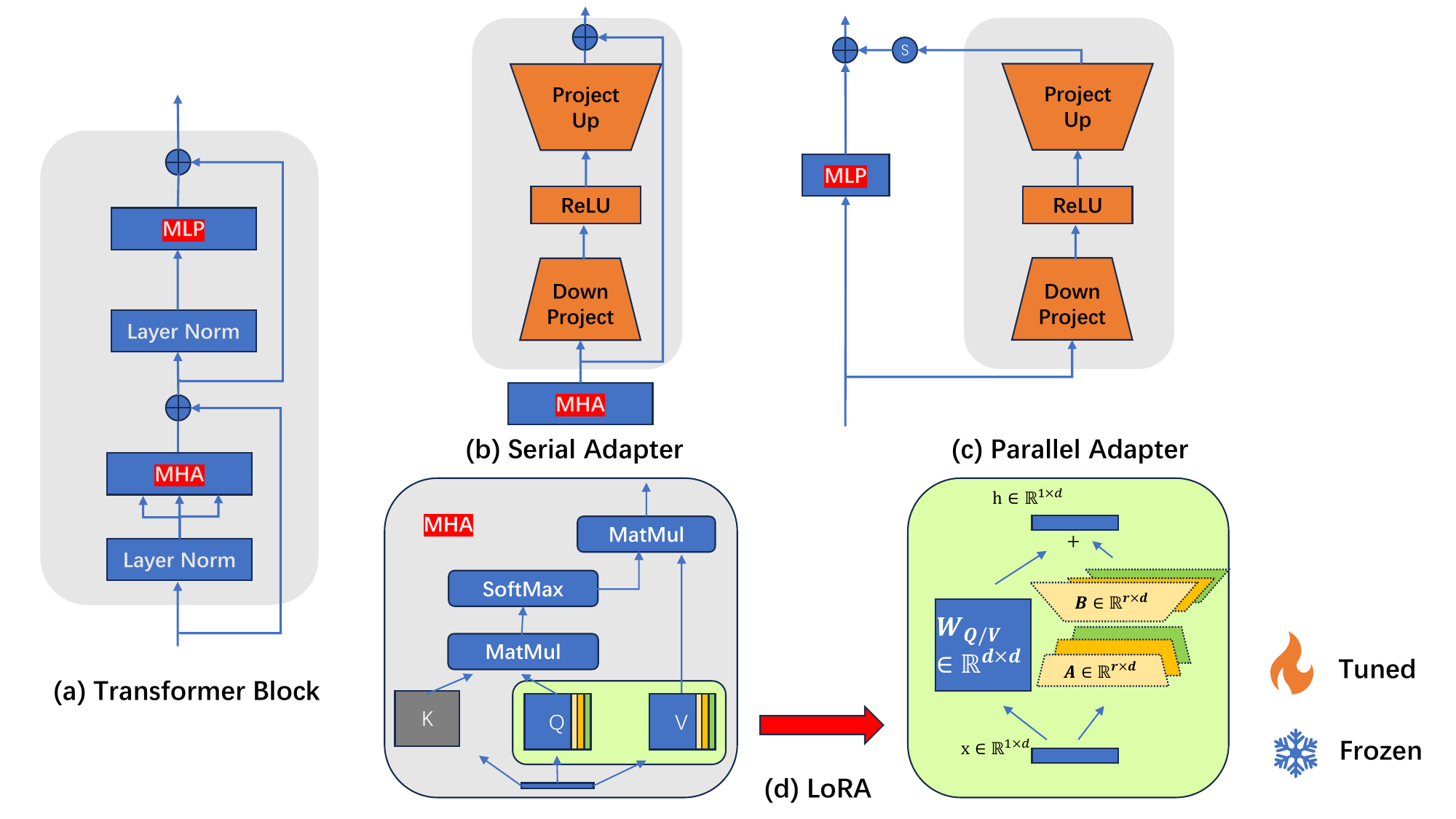}
    \caption{SelaFD: ViT fine-tuned architecture for HAR.}
    \label{fig:xx}
\end{figure}

\subsection{LoRA fine-tuning for weight space}

In order to efficiently transfer ViT models trained on natural images to radar-based TD signatures, we use LoRA fine-tuning, which is a classical parameter-efficient fine-tuning method that extracts basic coarse-grained features such as colors and shapes that are similar to those of natural images. A classical transformer block is shown in Fig. 2(a). Based on the premise assumption that the pre-trained large model weight changes are highly sparse and mainly play a role in lower intrinsic rank, we first employ LoRA fine-tuning for the Multi-Head Attention (MHA) part, with the structure shown in Fig. 2(d).

Specifically, we use the layer freezing technique to freeze the weights of the original model and greatly reduce the training parameters.In the attention layer, the pre-trained query and value projection matrices (denoted as \( \mathbf{W^Q} \) and \( \mathbf{W^V} \)), are fine-tuned with the addition of LoRA weights, which constrain their updates by employing a low-rank decomposition. This process can be mathematically represented as:

\begin{equation}
\mathbf{h} = \mathbf{W_0x} + \Delta \mathbf{Wx} = \mathbf{W_0x} + \mathbf{BAx},
\end{equation}
where \( \mathbf{x} \in \mathbb{R}^{1 \times d} \) and \( \mathbf{h} \in \mathbb{R}^{1 \times d} \) represent the input and output features, respectively. The weight change \( \Delta \mathbf{W} \in \mathbb{R}^{d \times d} \) of the pre-trained weight \( \mathbf{W}_0 \) is decomposed into two low-rank matrices, \( \mathbf{B} \in \mathbb{R}^{d \times r} \) and \( \mathbf{A} \in \mathbb{R}^{r \times d} \).The rank \( r \) of these matrices is significantly smaller than the model dimension \( d \).

\subsection{Serial-parallel adapter fine-tuning for feature space}
For fine-tuning general natural image tasks, the LoRA method effectively manages the weight space, rendering additional Adapter fine-tuning with superimposed feature spaces unnecessary. However, in the case of TD graphs, we introduce a novel serial-parallel combined adapter fine-tuning approach specifically designed for ambiguous categories to address the limitations of the LoRA method and enhanced fine-grained feature extraction.

In order to further enhance the feature representation and fine-grained feature extraction capability of the radar spectrum image, we add two adapters to the classical block. As shown in Fig. 2(b) and Fig. 2(c) in our approach, each transformer block is enhanced with two adapters. These adapters serve as bottleneck modules: they initially reduce the input dimensionality using a fully connected layer, apply a ReLU activation, and then project the output back to the original dimension. The first adapter operates serially after the MHA layer, featuring an internal skip-connection. The second adapter functions in parallel alongside the MLP layer, scaled by a factor \( s \). The computation for each globally adapted transformer block can be expressed as:
\begin{equation}
\mathbf{x}'_i = \text{Serial-Adapter}\left(\text{MHA}\left(\text{LN}(\mathbf{x}_{i-1})\right)\right) + \mathbf{x}_{i-1},
\end{equation}
\begin{equation}
\mathbf{x}_i = \text{MLP}\left(\text{LN}(\mathbf{x}'_i)\right) + \mathbf{s} \cdot \text{Parallel-Adapter}\left(\text{LN}(\mathbf{x}'_i)\right) + \mathbf{x}'_i,
\end{equation}
where \( \mathbf{x}_{i-1} \) and \( \mathbf{x}_i \) are the output of the \((i-1)\)-th and \(i\)-th transformer block. And the \text{LN} is the layer normalization to accelerate the training of neural networks, reduce the problems of gradient vanishing and gradient explosion, and improve the generalization performance of the network.

\section{EXPERIMENTS}
\label{sec:experiments}

\subsection{Datasets}
\label{ssec:dataset}

To investigate the generalizability of the proposed methodology, we utilized a public dataset collected by the University of Glasgow (UOG) at six different sites \cite{fioranelli2019radar}. The experiment included six specific activities: 1) walking, 2) sitting, 3) standing, 4) drinking, 5) picking up an object from the floor, and 6) falling. A total of 1,753 samples were used for the experiment.

To analyze the motion characteristics associated with six specific activities, we acquire Time-Doppler (TD) spectrograms using short time Fourier Transform (STFT) technique on the continuous waveform (CW) radar data. The resulting TD spectrograms effectively illustrate the temporal evolution of the Doppler frequencies, clearly presenting the micro-Doppler features associated with each activity, as shown in Fig. 3. This approach allows us to accurately distinguish between the six activities based on their unique motion patterns.

\begin{figure}[t]
    \centering
    \begin{minipage}[b]{0.25\linewidth}
        \centering
        \includegraphics[width=\linewidth]{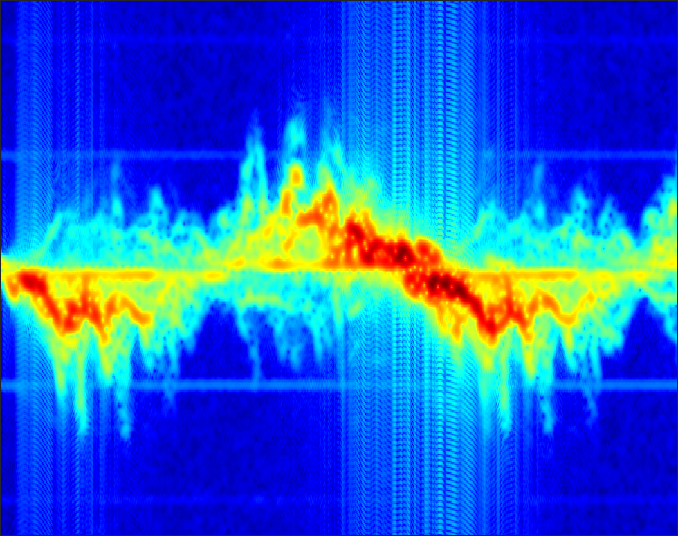}
        \centerline{(1) Walking}
        \label{fig:walk}
    \end{minipage}
    \hfill
    \begin{minipage}[b]{0.25\linewidth}
        \centering
        \includegraphics[width=\linewidth]{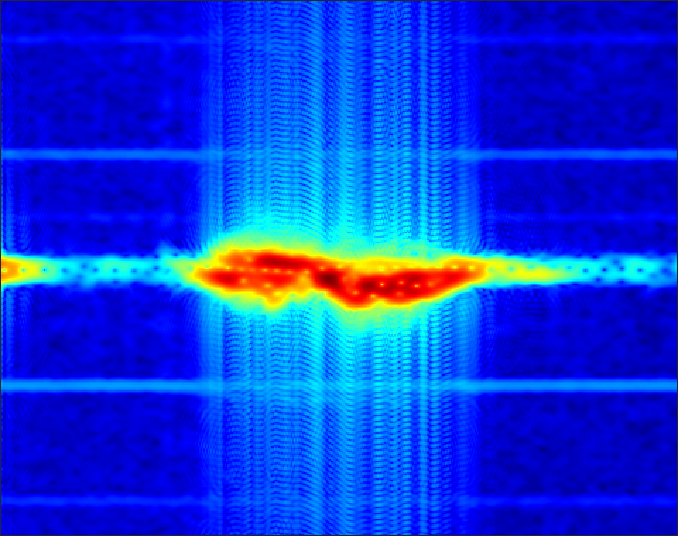}
        \centerline{(2) Sitting}
        \label{fig:sit}
    \end{minipage}
    \hfill
    \begin{minipage}[b]{0.25\linewidth}
        \centering
        \includegraphics[width=\linewidth]{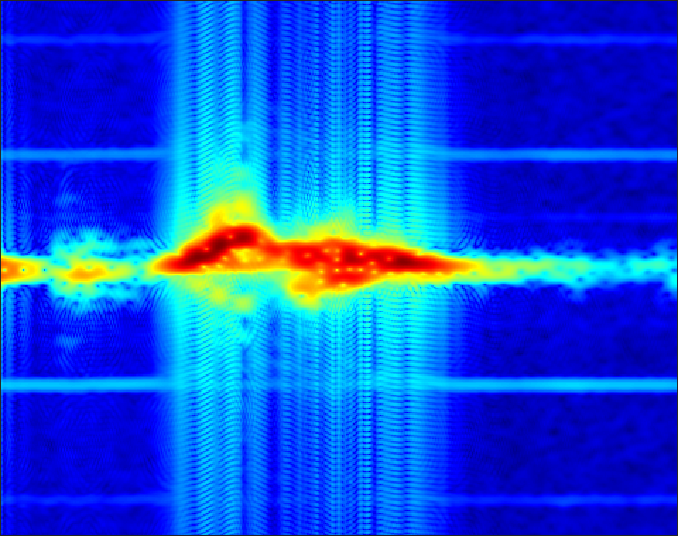}
        \centerline{(3) Standing}
        \label{fig:stand}
    \end{minipage}
    
    \vspace{0.1cm}
    
    \begin{minipage}[b]{0.25\linewidth}
        \centering
        \includegraphics[width=\linewidth]{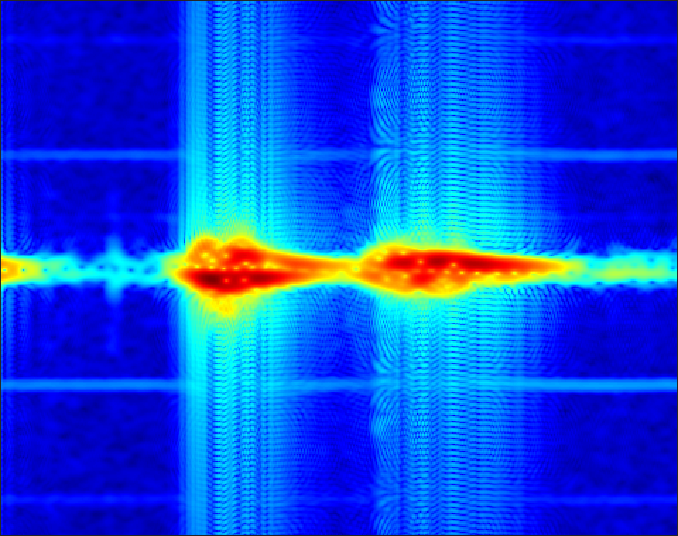}
        \centerline{(4) Drinking}
        \label{fig:drink}
    \end{minipage}
    \hfill
    \begin{minipage}[b]{0.25\linewidth}
        \centering
        \includegraphics[width=\linewidth]{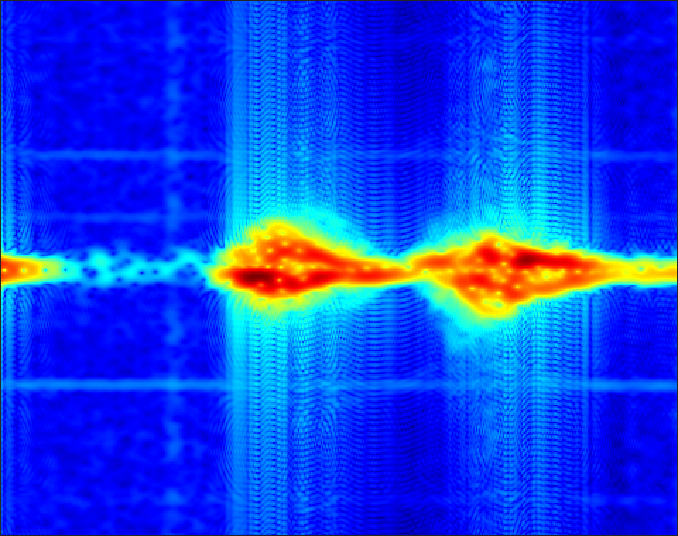}
        \centerline{(5) Picking up}
        \label{fig:pick}
    \end{minipage}
    \hfill
    \begin{minipage}[b]{0.25\linewidth}
        \centering
        \includegraphics[width=\linewidth]{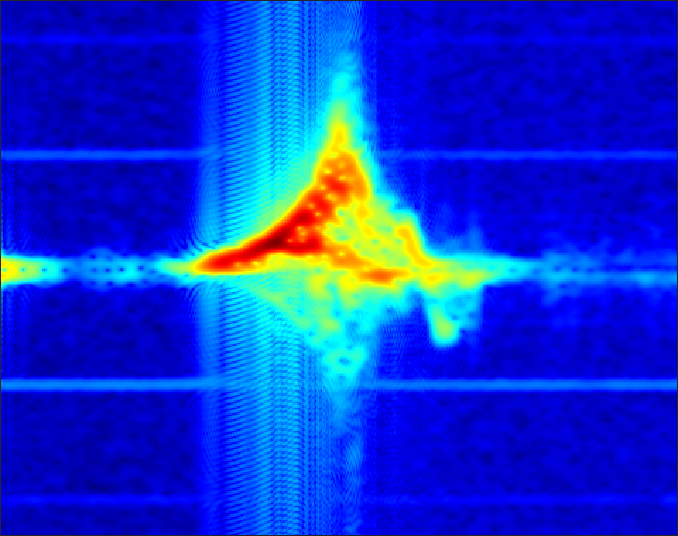}
        \centerline{(6) Falling}
        \label{fig:fall}
    \end{minipage}
    
    \caption{TD maps for the six specific activities.}
    \label{fig:all_images}
    
\end{figure}

\begin{figure}[t]
    \centering
    \begin{minipage}[b]{0.25\linewidth}
        \centering
        \includegraphics[width=\linewidth]{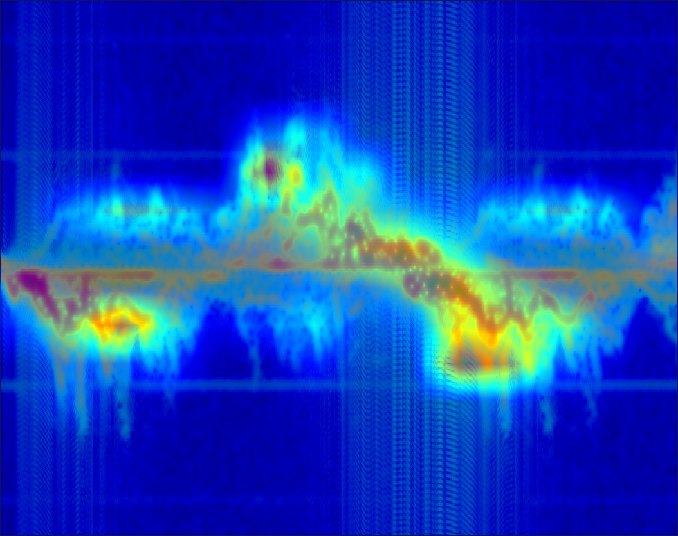}
        \centerline{(1) Walking}
        \label{fig:walk_a}
    \end{minipage}
    \hfill
    \begin{minipage}[b]{0.25\linewidth}
        \centering
        \includegraphics[width=\linewidth]{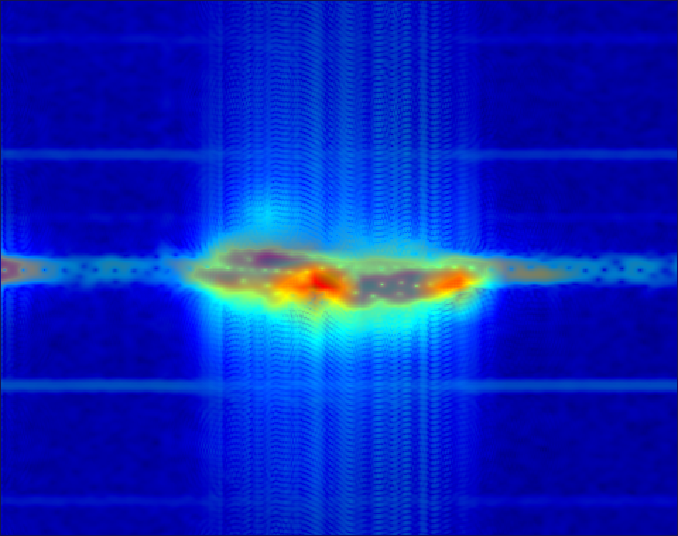}
        \centerline{(2) Sitting}
        \label{fig:sit_a}
    \end{minipage}
    \hfill
    \begin{minipage}[b]{0.25\linewidth}
        \centering
        \includegraphics[width=\linewidth]{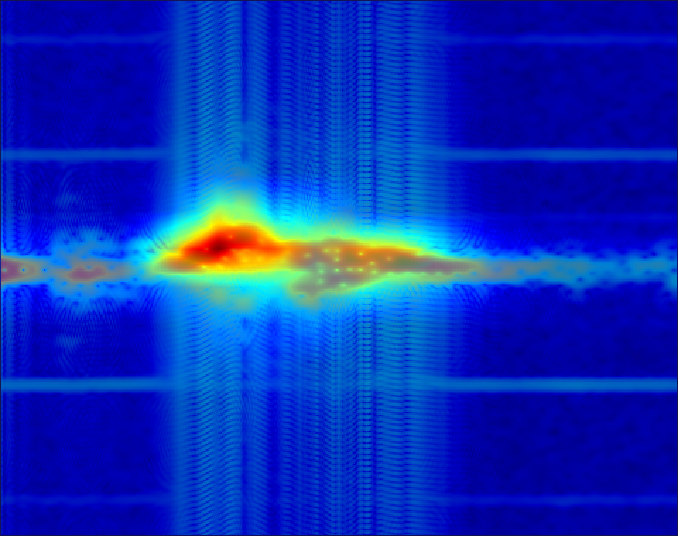}
        \centerline{(3) Standing}
        \label{fig:stand_a}
    \end{minipage}
    
    \vspace{0.1cm}
    
    \begin{minipage}[b]{0.25\linewidth}
        \centering
        \includegraphics[width=\linewidth]{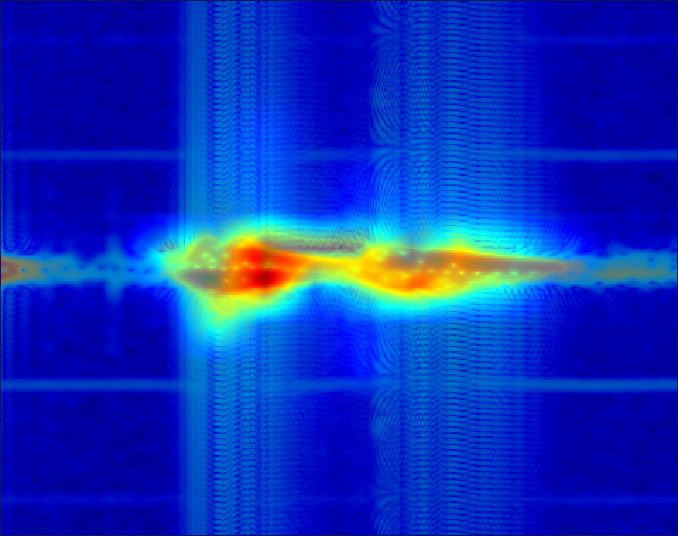}
        \centerline{(4) Drinking}
        \label{fig:drink_a}
    \end{minipage}
    \hfill
    \begin{minipage}[b]{0.25\linewidth}
        \centering
        \includegraphics[width=\linewidth]{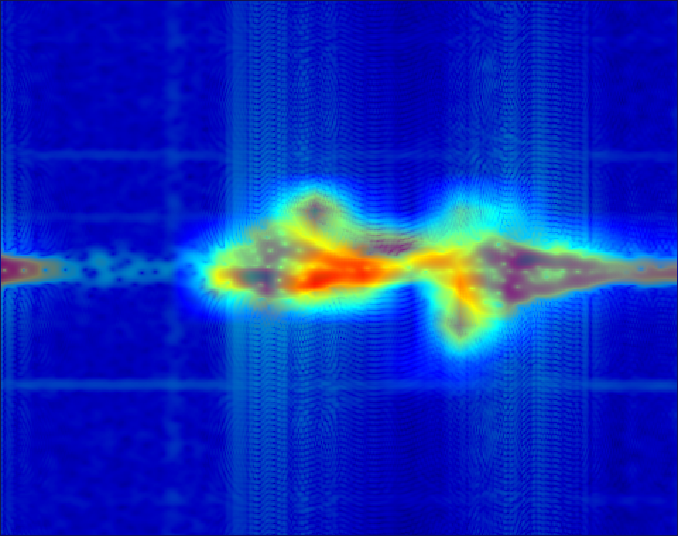}
        \centerline{(5) Picking up}
        \label{fig:pick_a}
    \end{minipage}
    \hfill
    \begin{minipage}[b]{0.25\linewidth}
        \centering
        \includegraphics[width=\linewidth]{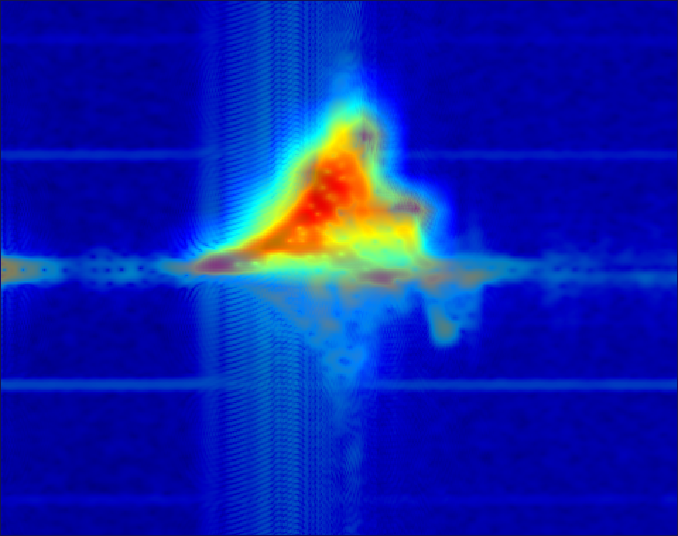}
        \centerline{(6) Falling}
        \label{fig:fall_a}
    \end{minipage}
    
    \caption{TD attention maps for the six specific activities.}
    \label{fig:all_attn_images}
\end{figure}

\begin{figure*}[t]
    \centering
    \begin{minipage}[b]{0.3\linewidth} 
        \centering
        \includegraphics[width=\linewidth]{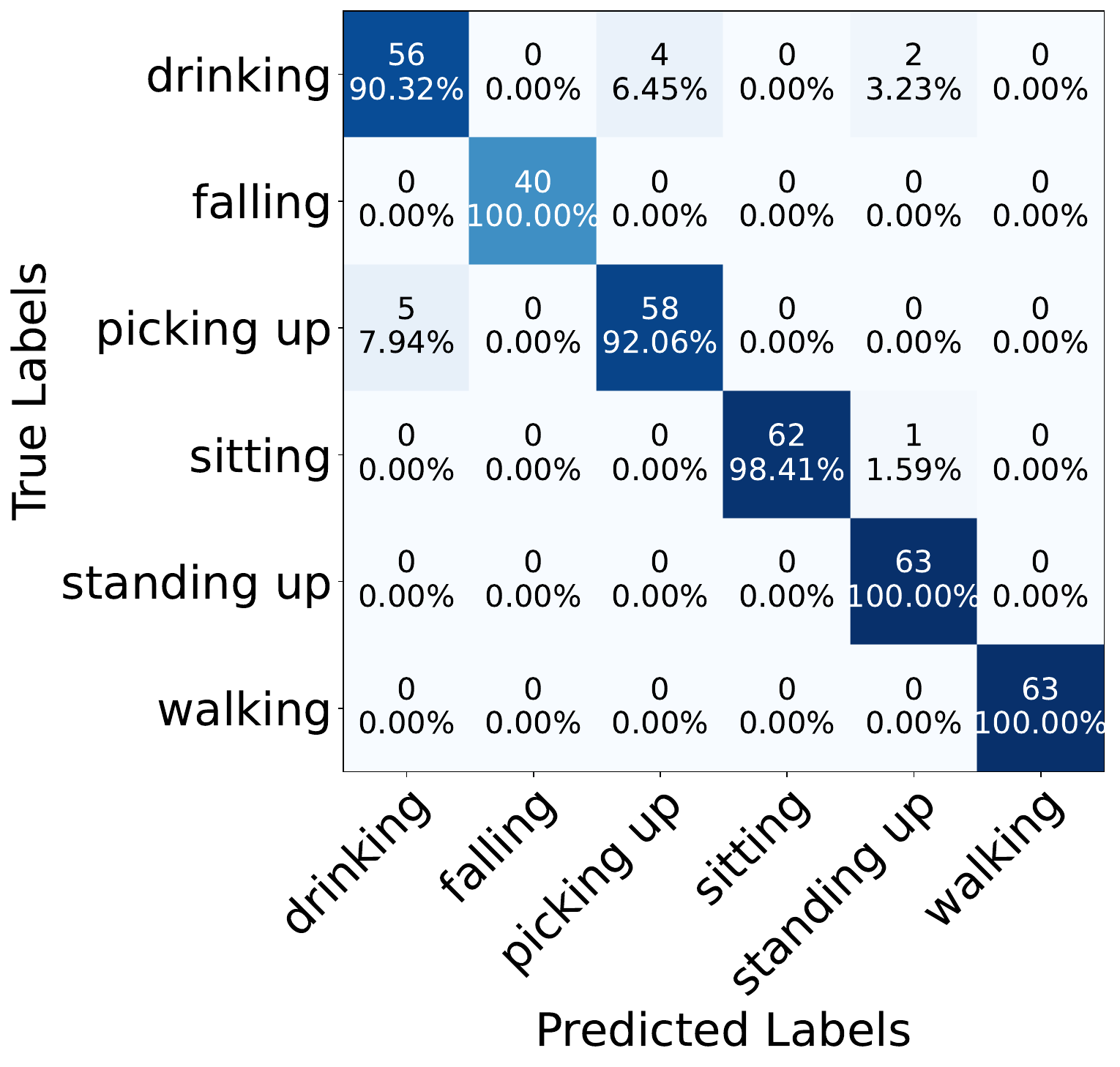}
        \centerline{(1) SelaFD}
        \label{fig:selafd-con}
    \end{minipage}
    \hfill
    \begin{minipage}[b]{0.3\linewidth}
        \centering
        \includegraphics[width=\linewidth]{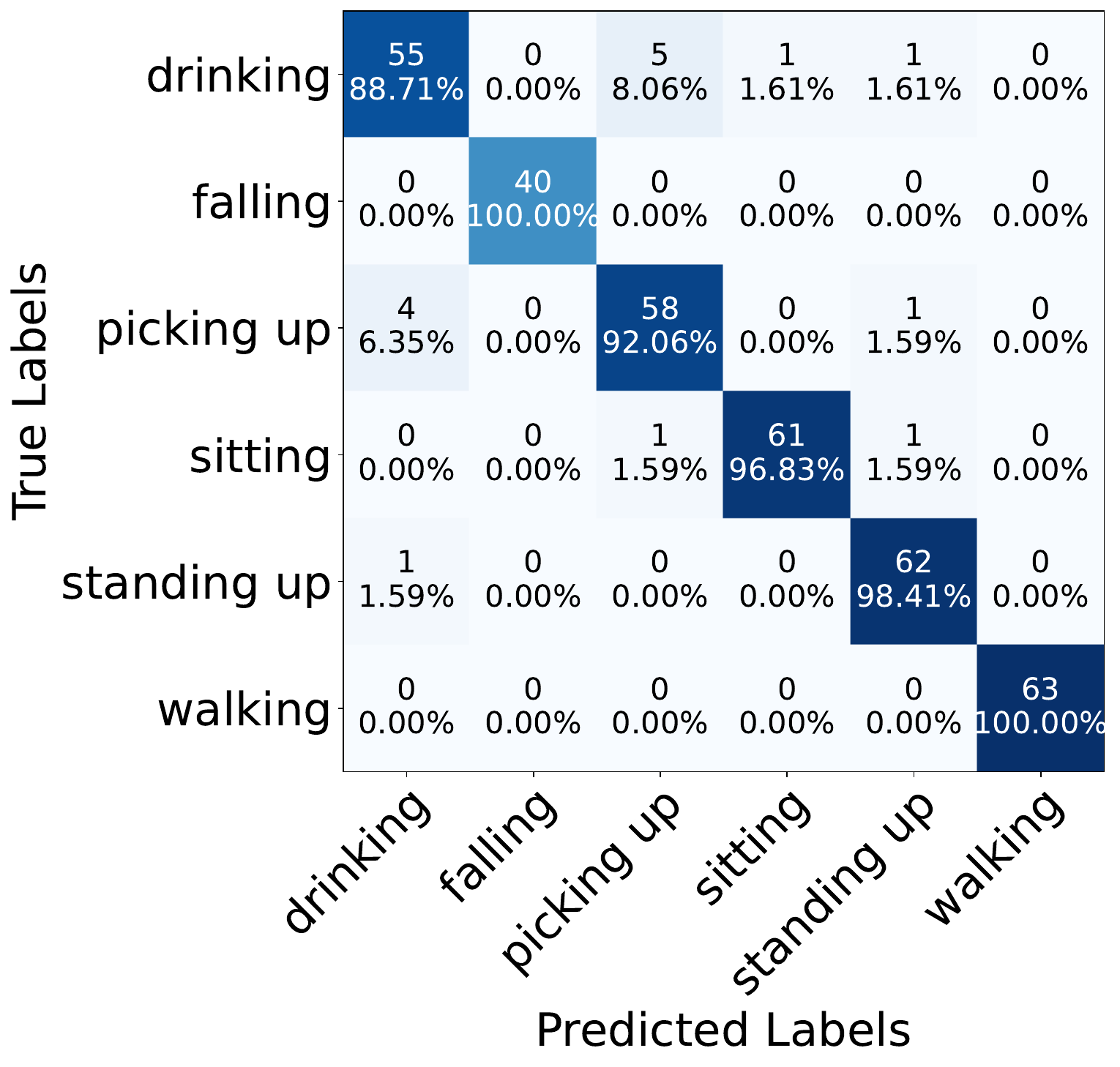}
        \centerline{(2) Adapter}
        \label{fig:ada}
    \end{minipage}
    \hfill
    \begin{minipage}[b]{0.3\linewidth}
        \centering
        \includegraphics[width=\linewidth]{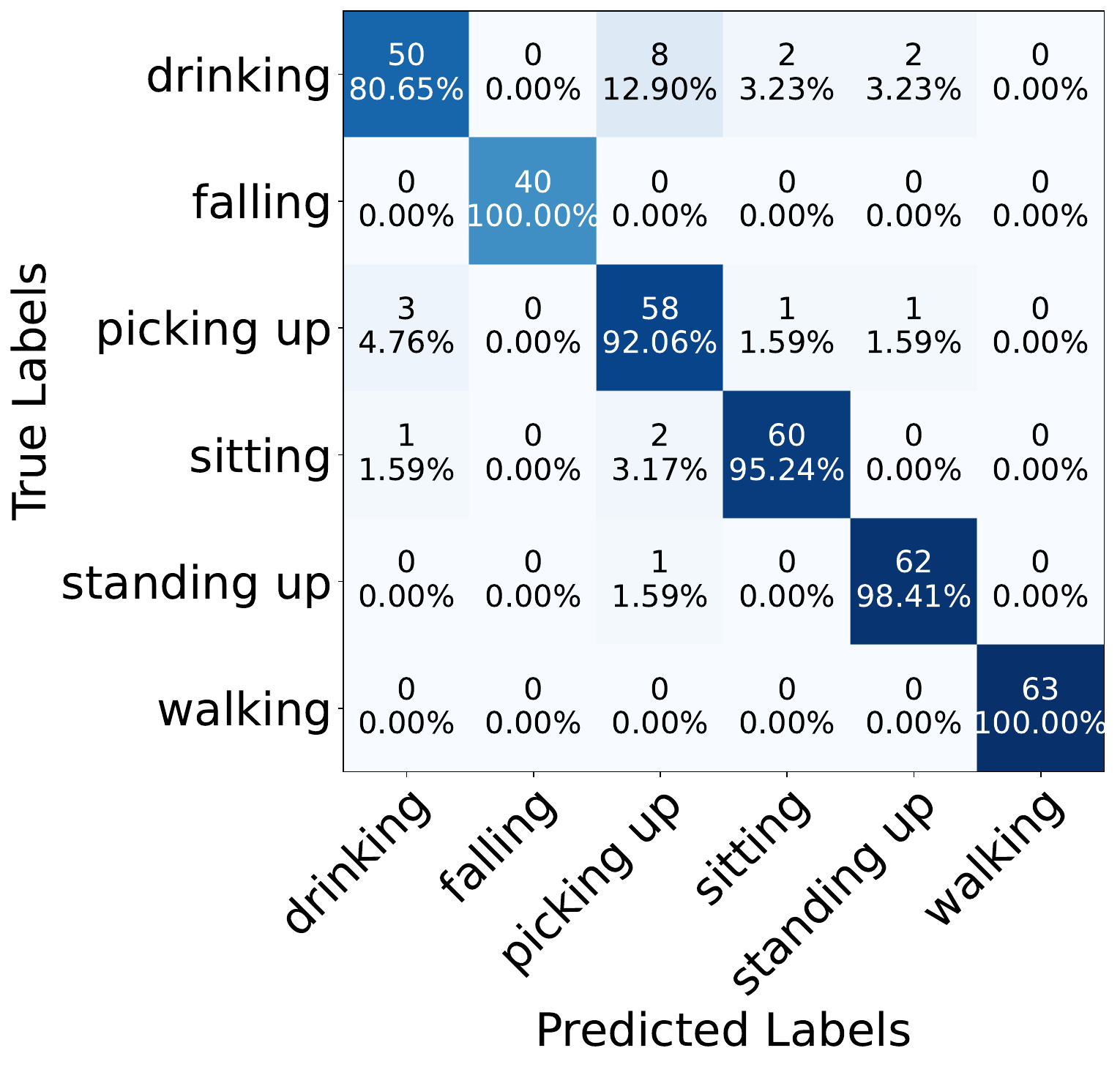}
        \centerline{(3) LoRA}
        \label{fig:lora}
    \end{minipage}
    \caption{Confusion matrix maps for the six specific activities using SelaFD, Adapter and LoRA methods.}
    \label{fig:con_mat}
\end{figure*}

\subsection{Implementation details}
\label{ssec:details}

We employ ViT-B/16, pretrained on ImageNet-21k, as our base model and conduct all experiments using PyTorch on NVIDIA Tesla A100 GPUs. The model generates a 768-dimensional feature vector for a 224×224 image. The bottleneck ratio of the adapters within the ViT block is 0.5, and the scaling factor \(s\) in Equation 3 is configured to 0.2. During training, we utilize the Adam optimizer with a learning rate of 0.0001, a batch size of 128, and employ the CosineAnnealingLR scheduling technique. The model is trained for 200 epochs.

\subsection{Algorithmic effect}
\label{ssec:effect}
We performed the HAR task according to the proposed joint fine-tuning of weight space and feature space as shown in Fig. 2. In the attention maps in Fig. 4, we can clearly see that the model focuses well on the core part of the TD maps. In Table I, we compare the classification accuracy of various methods based on CNN architectures such as ResNet and LSTM, or a combination of ViT and CNN, with our SelaFD method on the public UOG dataset over recent years.

\begin{table}[ht]
\centering
\caption{Performance Comparison of the Proposed Method with Recent Approaches Applied to the Public UOG Dataset for HAR.}
\label{table:methods}
\begin{tabular}{cc}
\toprule
\textbf{Method} & \textbf{Accuracy} \\
\midrule
WRGAN-GP+DCNN \cite{qu2022human} & 92.30\%   \\
CWT+RD-CNN \cite{kim2022radar} & 95.71\%   \\
CA-CFAR+PointNet \cite{fioranelli2022benchmarking} & 88.00\%   \\
TDSP+3D-PointNet \cite{ding2023sparsity} & 92.16\%   \\
DRDSP+4D-PointNet \cite{ding2023sparsity} & 95.69\%   \\
LH-ViT \cite{huan2023lightweight} & 92.10\%  \\
\textbf{SelaFD (Ours)} & \textbf{96.61\%}  \\
\bottomrule
\end{tabular}
\end{table}

Compared to these recent approaches, we have the following characteristics:

1. Leading accuracy with 100\% fall detection rate, as shown in the confusion matrix diagram in Fig. 5;

    
2. Successfully improved the ability to transfer natural image knowledge to radar physical images by proposing an effective fine-tuning method for TD maps, this will be further illustrated and explained in ablation experiments section.

\subsection{Ablation experiments}
\label{ssec:ablation}
In order to illustrate the effectiveness of the model architecture we designed, we conducted a number of ablation and comparison experiments, including linear fine-tuning and full fine-tuning without the adapter and lora modules, LoRA fine-tuning without the adapter module, and adapter fine-tuning without the lora module.The results of the experiment are in Table II. It can be seen that both the Adapter fine-tuning with the weight-space lora module removed and the LoRA fine-tuning with the feature-space adapter module removed show a decrease in classification accuracy compared to the full SelaFD fine-tuning. And the linear fine-tuning and full-parameter fine-tuning are much less accurate than the above methods.

Furthermore, to further demonstrate the validity of our proposed method, we present confusion matrix maps for the Adapter and LoRA methods, as illustrated in Fig. 5. When compared to the confusion matrix of the SelaFD method, it becomes evident that the SelaFD method significantly enhances the classification accuracy for the drinking category and reduces its confusability with the picking up category, which has been a particularly difficult category to differentiate between representatives of the drinking and picking up categories. This just proves that the adapter module we designed has the ability to extract fine-grained features for distinguishing confusable categories. Notably, this improvement is achieved without compromising the classification performance for other categories.

\begin{table}[ht]
\centering
\caption{SelaFD ablation experiments with ViT-B/16 on the Public UOG Dataset for HAR.}
\label{table:ablation}
\begin{tabular}{cc}
\toprule
\textbf{Method} & \textbf{Accuracy} \\
\midrule
w/o adapter  & 94.07\%   \\
w/o lora & 95.34\%   \\
w/o adapter\&lora (Linear)  & 84.75\%   \\
w/o adapter\&lora (Full) & 84.18\%   \\
\textbf{SelaFD (Ours)} & \textbf{96.61\%}  \\
\bottomrule
\end{tabular}
\end{table}

\section{Conclusions}
\label{sec:conclusions}

To the best of our knowledge, this work is the first to propose the application of ViT fine-tuning and transfer learning to the field of HAR, specifically using TD maps for HAR. In this study, we propose a method of joint fine-tuning in both weight space and feature space for TD maps, a type of special radar-based Time-Doppler signatures. This approach significantly improves the accuracy of HAR. 

In summary, this research pioneers the integration of radar signals with cutting-edge ViT architecture in the field of computer vision. By successfully adapting ViT for the fine-grained feature extraction of Time-Doppler maps, we not only demonstrate the potential of radar-based data in HAR but also highlight the versatility of ViT in handling non-traditional image modalities. Our findings establish a robust framework that can be leveraged for future advancements in processing radar signal images, paving the way for the development of more sophisticated and accurate HAR systems.

\vfill\pagebreak

\bibliographystyle{IEEEbib}
\bibliography{main}

\end{document}